\documentclass[10pt,twocolumn,letterpaper]{article}

\usepackage{cvpr}
\usepackage{times}
\usepackage{epsfig}
\usepackage{graphicx}
\usepackage{amsmath}
\usepackage{amssymb}
\usepackage{bm}
\usepackage{multirow}
\usepackage{indentfirst}
\usepackage{color}
\usepackage{eso-pic}

\usepackage[pagebackref=true,breaklinks=true,letterpaper=true,colorlinks,bookmarks=false]{hyperref}

\newcommand{\tabincell}[2]{\begin{tabular}{@{}#1@{}}#2\end{tabular}}

\newcommand\blfootnote[1]{%
  \begingroup
  \renewcommand\thefootnote{}\footnote{#1}%
  \addtocounter{footnote}{-1}%
  \endgroup
}

\cvprfinalcopy 

\def\cvprPaperID{619} 

\ifcvprfinal\fi
\begin{document}

\title{Context-aware Biaffine Localizing Network for Temporal Sentence Grounding}

\author{
Daizong Liu\textsuperscript{1*} \ \
Xiaoye Qu\textsuperscript{2*} \ \
Jianfeng Dong\textsuperscript{3} \ \
Pan Zhou\textsuperscript{1$\dagger$} \ \
Yu Cheng\textsuperscript{4} \ \
Wei Wei\textsuperscript{1} \\
Zichuan Xu\textsuperscript{5} \ \
Yulai Xie\textsuperscript{1}
\\
\textsuperscript{1}Huazhong University of Science and Technology  \quad \textsuperscript{2}Huawei Cloud \\
\textsuperscript{3}Zhejiang Gongshang University
\quad \textsuperscript{4}Microsoft AI\\
\textsuperscript{5}Dalian University of Technology
\\
{\tt\small \{dzliu, panzhou, weiw, ylxie\}@hust.edu.cn  \quad quxiaoye@huawei.com} \\ {\tt\small dongjf24@gmail.com \quad yu.cheng@microsoft.com \quad z.xu@dlut.edu.cn}
}

\maketitle
\thispagestyle{empty}
\begin{abstract}
\vspace{-11pt}
This paper addresses the problem of temporal sentence grounding (TSG), which aims to identify the temporal boundary of a specific segment from an untrimmed video by a sentence query. Previous works either compare pre-defined candidate segments with the query and select the best one by ranking, or directly regress the boundary timestamps of the target segment. 
In this paper, we propose a novel localization framework that scores all pairs of start and end indices within the video simultaneously with a biaffine mechanism.
In particular, we present a \textbf{C}ontext-aware \textbf{B}iaffine \textbf{L}ocalizing \textbf{N}etwork (CBLN) which incorporates both local and global contexts into features of each start/end position for biaffine-based localization. 
The local contexts from the adjacent frames help distinguish the visually similar appearance, and the global contexts from the entire video contribute to reasoning the temporal relation. 
Besides, we also develop a multi-modal self-attention module to provide fine-grained query-guided video representation for this biaffine strategy.
Extensive experiments show that our CBLN significantly outperforms state-of-the-arts on three public datasets (ActivityNet  Captions, TACoS, and Charades-STA), demonstrating the effectiveness of the proposed localization framework.
The code is available at  \href{https://github.com/liudaizong/CBLN}{https://github.com/liudaizong/CBLN}.
\vspace{-10pt}
\end{abstract}
\vspace{-1mm}
\blfootnote{
\textsuperscript{$*$}Equal contributions. ~~~~\textsuperscript{$\dagger$}Corresponding author.}
\section{Introduction}
\begin{figure}[t]
\vspace{-10pt}
\centering
\includegraphics[width=0.475\textwidth]{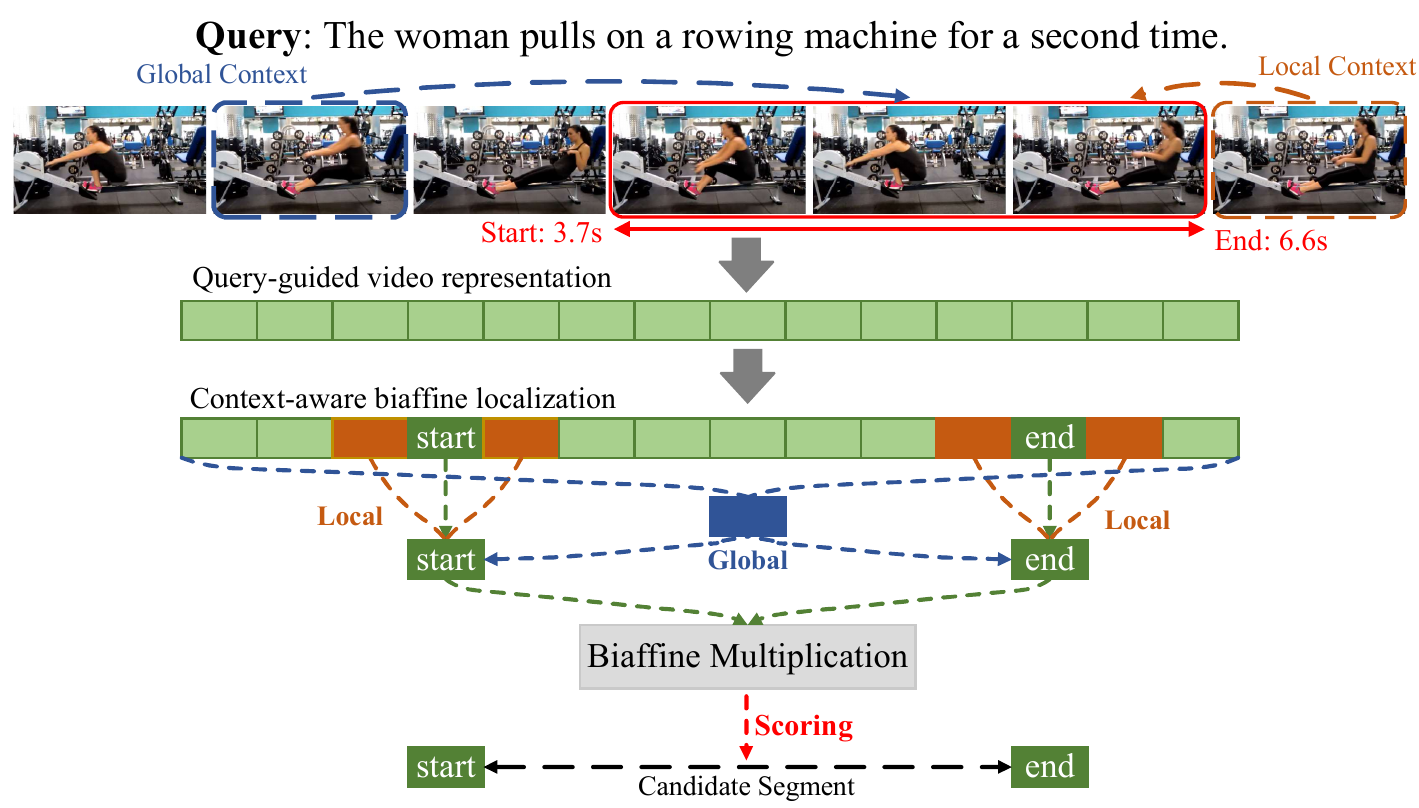}
\caption{An illustrative example of temporal sentence grounding task. From a new perspective, we propose a biaffine-based localization method that scores all pairs of start and end frames simultaneously with both local and global contexts.}
\label{fig:introduction}
\vspace{-10pt}
\end{figure}

\vspace{-1mm}
Video understanding is a fundamental task in computer vision and has drawn increasing attention over the last years due to its various applications in video event detection \cite{videvent}, video summarization \cite{song2015tvsum,chu2015video,liu2020violin}, video captioning \cite{jiang2018recurrent,chen2020learning,li2020hero} and temporal action localization \cite{shou2016temporal,zhao2017temporal}, etc.
Recently, temporal sentence grounding (TSG) \cite{gao2017tall,anne2017localizing} has been proposed as an important yet challenging task. This task requires automatically determining the start and end timestamps of a target segment in an untrimmed video that contains an activity semantically corresponding to a given sentence description, as shown in Figure \ref{fig:introduction}. It is substantially more challenging as it needs to not only model the complex multi-modal interactions among vision and language features, but also capture complicated context information for their semantics alignment.

Most previous methods \cite{anne2017localizing,gao2017tall,ge2019mac,liu2018attentive,chen2018temporally,zhang2019cross,liu2018cross,yuan2019semantic} tackle TSG task following a multi-modal matching architecture, which generates multiple candidate proposals of different time intervals and ranks them according to their similarities with the sentence query. These methods severely rely on the quality of proposals, and break the intrinsic temporal structure and global context of videos.
Recently, several works \cite{rodriguez2020proposal,yuan2019find,chenrethinking,wang2019language,he2019read,mun2020local,zeng2020dense} directly regress the temporal locations of the target segment. Specifically, they either regress the start/end timestamps based on the entire video representation \cite{yuan2019find,mun2020local}, or predict at each frame to determine whether this frame is a start or end boundary \cite{rodriguez2020proposal,zeng2020dense,chenrethinking}. 
However, in these methods, start and end features are never jointly considered. Given a video with two target segments that have the same starting action (\textit{open the door}) but end with different actions (\textit{go in}, \textit{go out}), predicting the start point independently may lead to timestamp confusion. Moreover, the ending point also tends to be inaccurate if it is predicted conditioned on the wrong start time.

Different from the aforementioned frameworks, we address the TSG from a new perspective:  
we reformulate this task by scoring all pairs of start and end indices simultaneously with a biaffine mechanism which interacts characteristics of each pair of possible start and end frames. 
This biaffine-based architecture is inspired by the dependency parsing task \cite{dozat2017deep} in natural language processing, in which the system predicts a dependency head for each child token and assigns a relation to the head-child pairs. 
However, there are two main obstacles when the biaffine mechanism used in dependency parsing is applied to TSG task. First, different from adjacent words in a sentence which carry different meaning, video is continuous and the adjacent frames naturally contain visually similar appearance.
As shown in Figure \ref{fig:introduction}, the adjacent frames near the segment boundary possess the similar semantics on ``woman" and ``pulls". Thus, it is difficult to distinguish the specific boundary from adjacent frames without referring to these adjacent features as local context. 
Second, causalities between word pairs are usually indirect and can be far apart, but in videos, events in different intervals are directly correlated and rely on the whole contents to reason the precise semantics.
For example, in Figure \ref{fig:introduction}, without perceiving ``a second time" from a global perspective, the first and second time of ``pull on a rowing machine" possess similar semantic boundaries but totally different temporal indices. Therefore, the causal relations between the video events which act as global context are essential for understanding the segment. 

Based on the above considerations, we propose a novel \textbf{C}ontext-aware \textbf{B}iaffine \textbf{L}ocalizing \textbf{N}etwork (CBLN), for temporal sentence grounding. Specifically, we develop a multi-context biaffine localization (MCBL) module which aggregates both local and global contexts to enrich the information of each frame representation. For each frame, we span the entire video features with different window sizes to get multi-scale video events as global contexts, and extract different numbers of adjacent frame features as multi-scale local contexts. The multi-scale local and global contexts are then inter-modulated to produce more adaptive contexts, which serve as the input representation for further biaffine localization by concatenating with frame-wise feature.
At last, we obtain the output scores from the biaffine model to identify the similarities of all possible start-end pairs according to the semantics of sentence query. 
Besides, to provide fine-grained query-guided video representation for above biaffine localization, we also develop a multi-modal self attention (MMSA) module to sufficiently capture dependencies
among video frames under the guidance of the sentence description.
By jointly learning the overall model, our CBLN is able to localize query in video effectively.

Our main contributions are summarized as follows:
\begin{itemize}
\vspace{-9pt}
\item From a new perspective, we adopt biaffine mechanism to the TSG task. The biaffine-based architecture simultaneously scores all possible pairs of start and end frames for segment localization. Compared to previous methods, it gets rid of complicated proposal design and interacts both start-end timestamps effectively.  
\vspace{-8pt}
\item To alleviate the limitation of the biaffine localization, we further develop a multi-context biaffine localization module which utilizes multi-scale local and global contexts to enrich frame representations. 
\vspace{-8pt}
\item We conduct extensive experiments to validate the effectiveness of our proposed CBLN on three datasets (ActivityNet Captions, TACoS, and Charades-STA), and show that it significantly outperforms the state-of-the-arts by a large margin.
\end{itemize}

\section{Related Work}

\begin{figure*}[t]
\vspace{-10pt}
\centering
\includegraphics[width=1.0\textwidth]{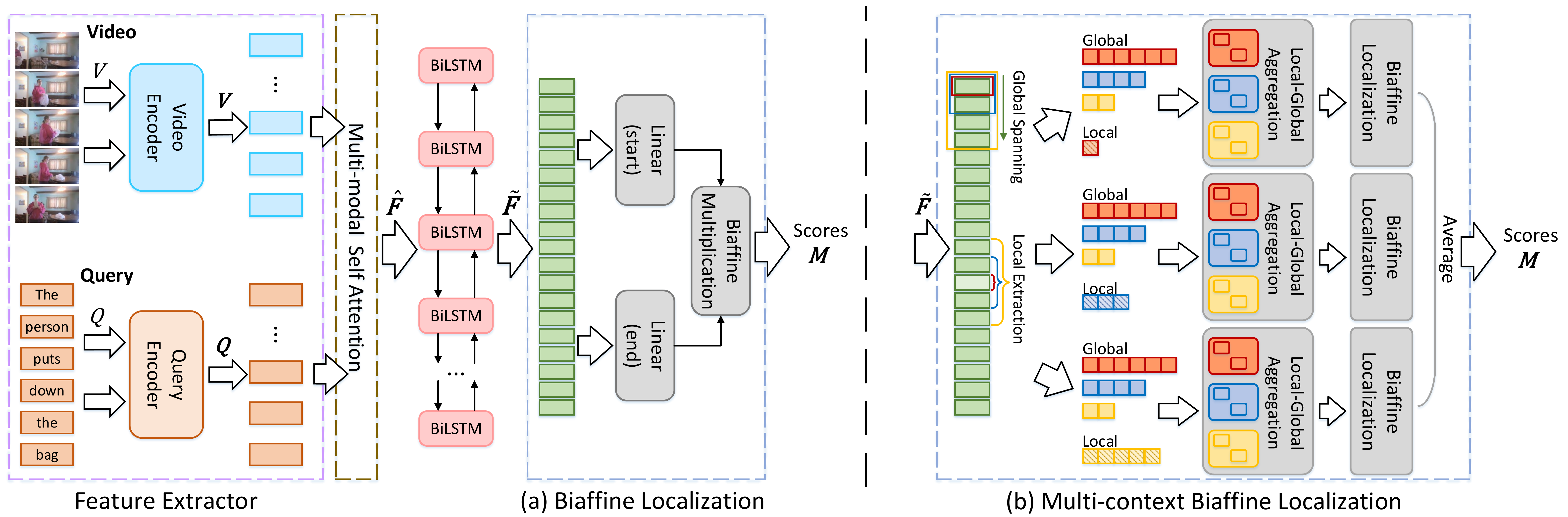}
\caption{An overview of the proposed architecture for TSG. Given a video and a sentence query, we first encode them by two feature extractors, and further interact them by a multi-modal self attention module. After obtaining query-guided video representations, we exploit a biaffine localization (Figure (a)) to score all possible segments simultaneously. Figure (b) is an improved version of Figure (a) and shows the structure of multi-context biaffine localization module which aggregates multi-scale local and global contexts.}
\label{fig:pipeline}
\vspace{-10pt}
\end{figure*}

\noindent \textbf{Temporal Sentence Grounding.}
Temporal sentence grounding (TSG) is a new task introduced recently \cite{gao2017tall,anne2017localizing,dong2021dual} that aims to retrieve video segments using language queries. 
The early works \cite{anne2017localizing,ge2019mac,liu2018attentive,zhang2019man,chen2018temporally,zhang2019cross,liu2018cross,yuan2019semantic,xu2019multilevel,liu2020jointly,liu2020reasoning} employ a multi-modal matching architecture that first generates segment proposals, and then ranks them according to the similarity between proposals and the query to select the best matching one. Some of them \cite{gao2017tall,anne2017localizing} propose
to apply the sliding windows to generate proposals and subsequently integrate the query with segment representations via a matrix operation. 
Instead of using the sliding windows, latest works \cite{wang2019temporally,zhang2019man,yuan2019semantic,zhang2019cross} directly integrate sentence information with each fine-grained video clip unit, and predict the scores of candidate segments by gradually merging the fusion feature sequence over time. Although those methods achieve promising performances, they are severely limited by the quality of proposals.

To overcome above drawback, recent works \cite{rodriguez2020proposal,yuan2019find,chenrethinking,wang2019language,he2019read,mun2020local,zeng2020dense} directly regress the temporal locations of the target segment.
Yuan \textit{et al.} \cite{yuan2019find} propose a co-attention based network to regress the start and end boundaries of the target segment. To improve the grounding with dense supervisions, Zeng \textit{et al.} \cite{zeng2020dense} regress the distances from each frame to the target start (end) frame. Chen \textit{et al.} \cite{chenrethinking} propose a graph based bottom-up framework to capture multi-level semantics and encode the plentiful scene relationships.

Different from aforementioned two types of methods, we give a new solution to address the TSG problem. Specifically, we regard each frame as start or end frame to build all possible candidate segments and score all of them simultaneously with a biaffine mechanism \cite{dozat2017deep}.
After that, we obtain the output scores for all segments and choose the best one corresponding to the highest value as the grounding result. Although 2D-TAN \cite{zhang2019learning} also scores all possible segments among the video, it directly max-pools the contained frame features to represent corresponding segment and lose discriminative frame-wise feature. Thus, compared to our detailed cross-modal interaction, this method fails to perform fine-grained interaction. Meanwhile, it captures the proposal-wise relations with convolution layers. Instead, our CBLN incorporates explicit local and global context to capture fine-grained frame-wise relations.

\noindent \textbf{Biaffine based Dependency Parsing.}
Biaffine mechanism is widely used in dependency parsing \cite{dozat2017deep,li2019self,yu2020named} which aims to build up a syntactic dependency tree for a given sentence. This task needs to capture all possible relations between the word pairs. Dozat \textit{et al.} \cite{dozat2017deep} are the first work to learn the long-dependency from a head word to a modifier word with a relation label by proposing biaffine. They utilize biaffine operation as a scoring algorithm to determine the syntactic of a phrase from one word to another. Yu \textit{et al.} \cite{yu2020named} further adapt biaffine to Named Entity Recognition \cite{finkel2009nested} by reformulating this task as the task of identifying start and end indices, as well as assigning a category to the span by these pairs. By treating input video as text passage, the biaffine mechanism is also applicable to TSG task in principle, because TSG aims to determine whether the segment from one frame to another is a full segment containing the activities described by the sentence query. However, the biaffine is not able to capture the adjacent contexts or correlate the global events, thus may lose local and global details about the scene meaning. In this paper, we enhance the biaffine mechanism by aggregating local-global information.

\section{Proposed Method}
Given an untrimmed video $\mathcal{V}$ and a sentence query $\mathcal{Q}$, we represent the video as $\mathcal{V}=\{v_t\}^{T}_{t=1}$ frame-by-frame, where $v_t$ is the $t$-th frame and $T$ is the number of total frames. Similarly, the query with $N$ words is denoted as $\mathcal{Q}=\{q_n\}^{N}_{n=1}$ word-by-word. Temporal sentence grounding (TSG) aims to localize a segment $(\tau_s, \tau_e)$ starting at timestamp $\tau_s$ and ending at timestamp $\tau_e$ in video $\mathcal{V}$, which corresponds to the same semantic as query $\mathcal{Q}$.

The key to our Context-aware Biaffine Localizing Network (CBLN) is that we score all pairs of start and end frames simultaneously with a biaffine mechanism by interacting the characteristics of the start-end pairs. As shown in Figure \ref{fig:pipeline}, we first utilize two encoders to extract both video and query features, and then introduce a multi-modal self attention (MMSA) to generate the fine-grained video features for localization. Subsequently, we exploit biaffine localization module to score the start and end frame pairs of all possible segments, as shown in Figure \ref{fig:pipeline} (a). In this way, we can get the scores for all candidate segments and choose the best one as the target segment. By enriching the query-guided video representation with local-global contexts, in Figure \ref{fig:pipeline} (b), we further propose a multi-context biaffine localization (MCBL) module for more precise grounding.

\begin{figure*}[t]
\vspace{-10pt}
\centering
\includegraphics[width=1.0\textwidth]{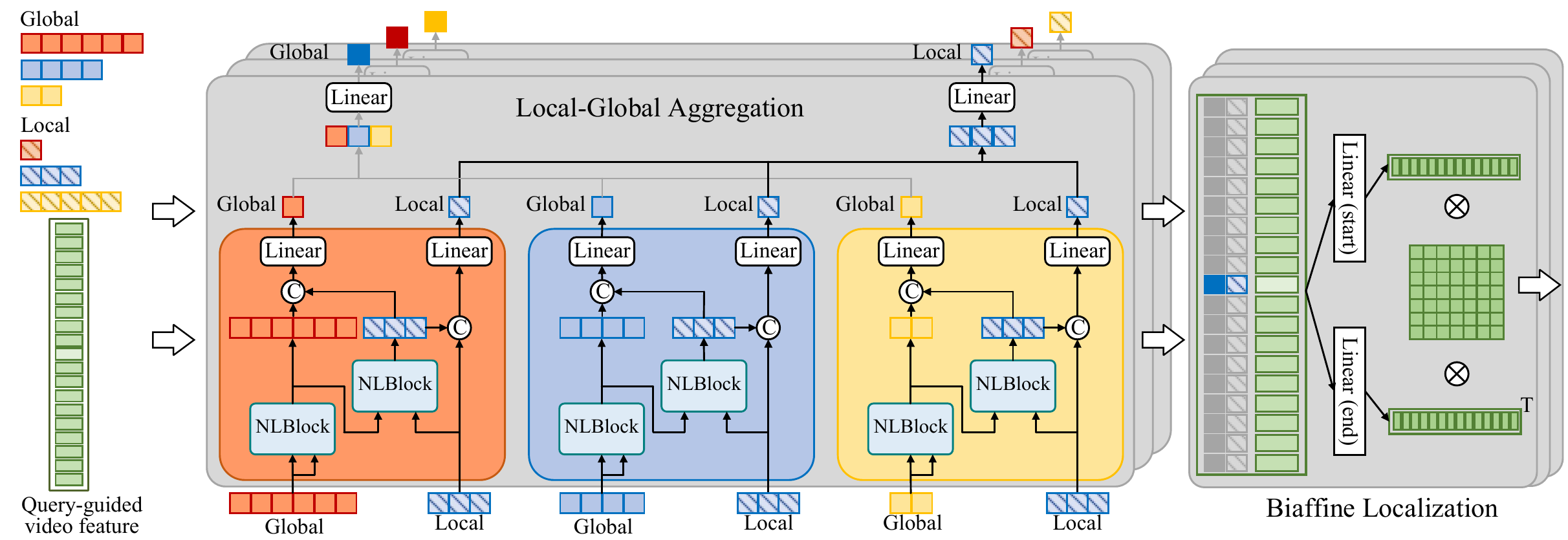}
\caption{An illustration of the process of aggregating both multi-scale local and global contexts to local-guided global and global-guided local contexts. These aggregated contexts are then concatenated with video representation for biaffine localization.}
\label{fig:biaffine}
\vspace{-10pt}
\end{figure*}

\subsection{Feature Extractor}
\noindent
\textbf{Video encoder.} For video encoding, we first extract the frame features by a pre-trained C3D network \cite{tran2015learning}, and then add a positional encoding \cite{vaswani2017attention} to take positional knowledge. Such position encoding plays a crucial role in distinguishing semantics at diverse temporal locations. Considering the sequential characteristic in videos, a bi-directional GRU \cite{chung2014empirical} is further utilized to incorporate the contextual information along time series. We denote the extracted video features as $\bm{V}=\{\bm{v}_t\}^{T}_{t=1} \in \mathbb{R}^{T \times D}$.

\noindent
\textbf{Query encoder.} For query encoding, we first extract the word embeddings by the Glove model \cite{pennington2014glove}. We also apply the positional encoding and bi-directional GRU to integrate the sequential information. The final feature of the input sentence query is denoted as $\bm{Q} = \{\bm{q}_n\}_{n=1}^N \in \mathbb{R}^{N \times D}$.

\subsection{Multi-Context Biaffine Localization}
\subsubsection{Biaffine Localization}
\indent 
Biaffine mechanism is widely used in dependency parsing \cite{dozat2017deep,li2019self,yu2020named} to assign scores to all possible spans in a sentence, where each span is defined by a pair of start and end words. As TSG can also be reformulated as the task of identifying the start and end frames of a specific segment among a given video, it is appropriate to adapt biaffine mechanism~\cite{dozat2017deep} to this task for scoring all possible segment candidates. In our CBLN, before interacting the features of start and end frames using the biaffine mechanism, we first propose a multi-modal self attention (illustrated in Section \ref{sec:mmsa}) to generate fine-grained query-guided video representation $\widehat{\bm{F}}=\{\widehat{\bm{f}}_t\}_{t=1}^T$, and then exploit a BiLSTM \cite{Mike1997} layer to further aggregate its sequential contexts as:
\begin{equation}
\setlength{\abovedisplayskip}{1pt}
\setlength{\belowdisplayskip}{1pt}
    \widetilde{\bm{F}} = \text{BiLSTM}(\widehat{\bm{F}}),
\label{eq:1}
\end{equation}
where $\widetilde{\bm{F}}=\{\widetilde{\bm{f}}_t\}_{t=1}^T$. 
Then, as the contexts of the start and end frames are different, we apply two separate linear layers to generate separate hidden representations for each pair of start and end frames:
\begin{equation}
    \bm{h}^s_p =  \widetilde{\bm{f}}_{p_s}\bm{W}^s + \bm{b}^s,
\label{eq:2}
\end{equation}
\begin{equation}
    \bm{h}^e_p =  \widetilde{\bm{f}}_{p_e}\bm{W}^e + \bm{b}^e,
\label{eq:3}
\end{equation}
where $p_s$ and $p_e$ are the start and end indexes of segment $p$, $\bm{W}^s, \bm{W}^e$, $\bm{b}^s, \bm{b}^e$ are the learnable parameters of two linear layers. Finally, we employ the biaffine operation over the start and end representations to generate scores $\bm{M}=\{\bm{M}_p\}_{p=1}^{T \times T} \in \mathbb{R}^{T\times T}$, which indicates the matching score of all segments. Details can be found in Figure \ref{fig:pipeline} (a). The scoring rule for each segment $p$ can be summarized as:
\begin{equation}
    \bm{M}_p = \sigma(\bm{h}^s_p \bm{U}^m (\bm{h}^e_p)^{\top} + (\bm{h}^s_p \oplus \bm{h}^e_p)\bm{W}^m + \bm{b}^m),
\label{eq:4}
\end{equation}
where $\bm{U}^m$ and $\bm{W}^m$ are learnable parameters, $\bm{b}^m$ is bias, and $\oplus$ denotes element-wise addition, $\sigma$ is the sigmoid function. After a sigmoid layer, $\bm{M}_p$ is the score of segment $p$, which indicates the probability of $p$ matched to the query. Experiments in next section demonstrate that biaffine mechanism achieves a superior performance in TSG task.

\subsubsection{Biaffine Localization with Local-Global Contexts}
Although biaffine localization module has a strong ability to address TSG, it measures each segment by only considering the features of its start and end frames. As a result, the context of segments can only be drawn from a limited extent of previous BiLSTM layer. 
To enrich the context information of the start and end frames, we integrate start/end representation with:  
1) Local contexts from adjacent frames. 
Since the adjacent frames near the segment boundary present similar visual appearance to the start/end frames, we aggregate the local contexts to distinguish them for more accurate boundaries grounding. 
2) Global contexts from the entire video. We also extract long-range contexts to reason the temporal relations between different events among the video. 
Moreover, as shown in Figure~\ref{fig:pipeline}~(b), the multi-scale local and global contexts are further aggregated. 
Finally, we concatenate each aggregated local-global contexts with corresponding start/end frame features for parallel multi-context biaffine localization.

\noindent \textbf{Local-Global contexts.}
Given a start/end frame $t$, we define two types of context features: ``local” features $\bm{R}^l_t$ extracted from adjacent frames, and ``global” features $\bm{R}^g_t$ spanned from the entire video. 
For ``local" features, since the local contexts cover several adjacent frames around the frame $t$,  we directly utilize a window of size $K^l$ on frame $t$ to extract features as follows:
\begin{equation}
\setlength{\abovedisplayskip}{1pt}
\setlength{\belowdisplayskip}{1pt}
    \bm{R}^l_t = \{\widetilde{\bm{f}}_{t-(K^l/2)},...,\widetilde{\bm{f}}_{t},...\widetilde{\bm{f}}_{t+(K^l/2)}\},
\end{equation}
where $\bm{R}^l_t \in \mathbb{R}^{K^l\times D}$. To capture more contextual details from local features, we generate local contexts of multiple scales by varying the number of $K^l=\{1,3,5\}$.

For ``global" features, as the global contexts refer to snippet-level features max-pooled from the long-range video frames, we define the spanning features $\bm{R}^g_t$ as a feature bank of snippet features with a window of size $K^g$ as:
\begin{equation}
\setlength{\abovedisplayskip}{1pt}
\setlength{\belowdisplayskip}{1pt}
    \bm{R}^g_{t,k} = \text{maxpool}(\widetilde{\bm{f}}_{(k-1)T/K^g},...,\widetilde{\bm{f}}_{kT/K^g-1}),
\end{equation}
\begin{equation}
    \bm{R}^g_t = \{ \bm{R}^g_{t,1}, \bm{R}^g_{t,2},...,\bm{R}^g_{t,k},...,\bm{R}^g_{t,(T/K^g)}\},
\end{equation}
where $\bm{R}^g_{t,k}\in\mathbb{R}^{1 \times D}$ is the max-pooled snippet features across a specific segment, and $\bm{R}^g_t \in \mathbb{R}^{T/K^g\times D}$ is the global contexts. We also generate multi-scale global contexts by varying $K^g=\{1,2,4\}$. We then ensemble these local-global contexts for further aggregation.


\noindent \textbf{Local-Global aggregation.} Since both local and global contexts need to be adaptive to the boundary frame at each position, we modulate them to generate local-guided global and global-guided local contexts.
In detail, we aggregate all scales of global features with each scale of local feature separately. In Figure \ref{fig:biaffine}, we demonstrate the details of multi-scale local-global aggregation. Given global feature of multiple scales and local feature of one specific scale, we first re-weight each local-global pair $(\bm{R}^l_t,\bm{R}^g_t)$ by:
\begin{equation}
    (\bm{R}^g_t)'= \text{NLBlock}(\bm{R}^g_t,\bm{R}^g_t),
\end{equation}
\begin{equation}
\setlength{\belowdisplayskip}{1pt}
    (\bm{R}^l_t)' = \text{NLBlock}((\bm{R}^g_t)',\bm{R}^l_t),
\end{equation}
where $\text{NLBlock}(\cdot)$ denotes a modified non-local block \cite{wu2019long} added with a layer normalization \cite{ba2016layer} and dropout layer \cite{srivastava2014dropout}. 
The first NLBlock is utilized to capture temporal relations between the pooled events. Since feature $(\bm{R}^g_t)'$ loses specific information on the interested frames, another NLBlock is designed to further model adaptive local contexts by reasoning the adjacent features with global events.
At last, we concatenate multi-level global/local contexts and project them through a linear layer to compute the final local-guided/global-guided global/local context for frame $t$. 

\begin{figure}[t]
\vspace{-10pt}
\centering
\includegraphics[width=0.40\textwidth]{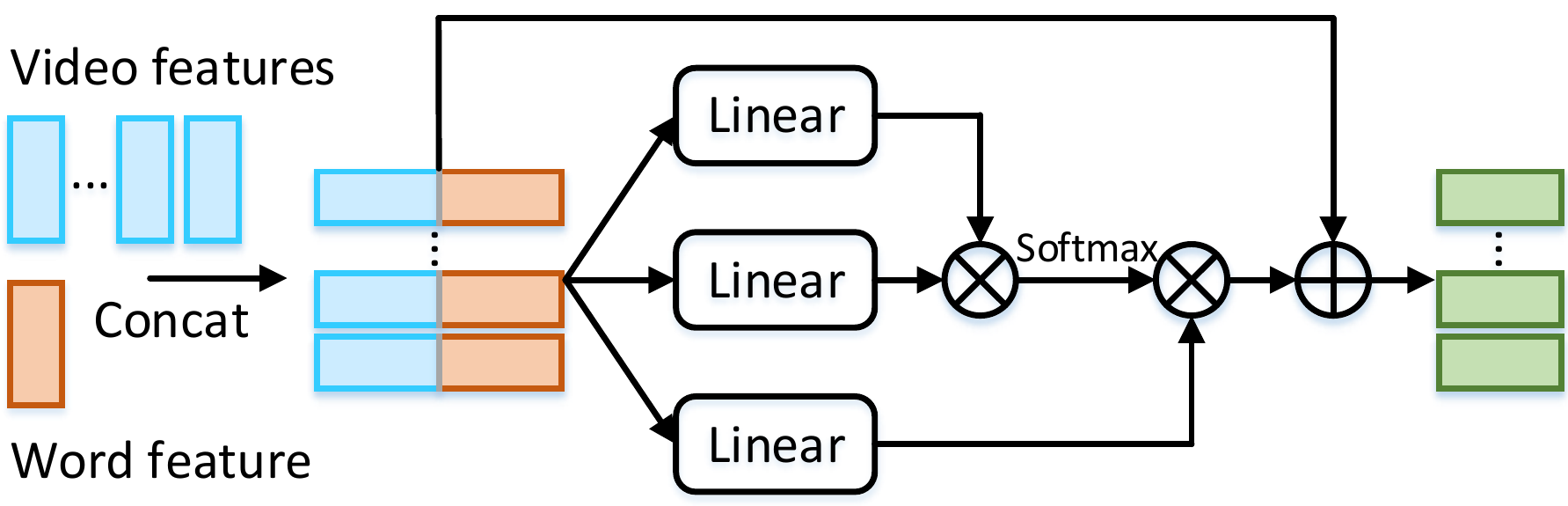}
\caption{The structure of multi-modal self attention on each word-video feature pair. Since query contains $N$ words, we employ $N$ such modules and average pool the corresponding outputs to generate fine-grained query-guided video representation.}
\label{fig:mmsa}
\vspace{-10pt}
\end{figure}

\noindent \textbf{Biaffine with multi-context.} After receiving the local-guided global and global-guided local contexts from each local-global aggregation module at frame $t$, we concatenate them to the corresponding frame feature $\widetilde{\bm{f}}_t$. Based on this feature, we can get a new context-aware query-guided video representation $(\widetilde{\bm{F}})'$, and then process it by Eq. (\ref{eq:2})(\ref{eq:3})(\ref{eq:4}) in biaffine localization module. At last, we average the results of size $T\times T$ from the outputs of multiple biaffine localization modules to produce the final scores $\bm{M}\in \mathbb{R}^{T\times T}$ for all segments, which is further followed by a sigmoid function. Each value on scores $\bm{M}$ represents the matching score between a segment with the queried sentence. 

\subsection{Multi-Modal Self Attention}
\label{sec:mmsa}
Building interaction between video and query is a crucial step to provide detailed query-guided video representation for localization. 
To achieve this goal, recent works \cite{chenrethinking,zhang2019cross,yuan2019find} interact each word with each frame by a co-attention mechanism. However, these methods only focus on frame-wise cross-modal matching and lack the interaction over long-range video frames under the query guidance, which is essential for consecutive semantics understanding. To learn a better query-guided video representation forming the input of our biaffine-based localization, as depicted in Figure \ref{fig:mmsa},     
we first concatenate word feature to each frame feature in a video and feed them into a multi-modal self attention module for long-range dependencies capturing.
Specifically, given a word $q_n$, we first construct a joint multi-modal feature by concatenating single word feature $\bm{q}_n$ to the whole video features $\bm{V}=\{\bm{v}_t\}^{T}_{t=1}$ as:
\begin{equation}
    \bm{F}_n = \{ \bm{f}_{nt}\}_{t=1}^T, \ \text{where} \ \bm{f}_{nt} = [\bm{v}_t;\bm{q}_n] \in \mathbb{R}^{1\times 2D}.
\end{equation}
The multi-modal self attention module then takes the multi-modal features $\bm{F}_n$ as input, and produces a set of query, key and value pair by linear transformations as
$\bm{l}^q_{nt} = \bm{f}_{nt}\bm{W}^q$, $\bm{l}^k_{nt} =  \bm{f}_{nt}\bm{W}^k$ and $\bm{l}^v_{nt} =  \bm{f}_{nt}\bm{W}^v$ at each frame $t$, where $\bm{W}^q, \bm{W}^k, \bm{W}^v$ are parameters to be learned. We compute the multi-modal self attentive feature $\widehat{\bm{l}}^v_{nt}$by:
\begin{equation}
    \widehat{\bm{l}}^v_{nt} = \sum^T_{t'=1} \alpha_{(nt,nt')} \bm{l}^v_{nt'}, \alpha_{(nt,nt')} = \text{Softmax}(\bm{l}^q_{nt} (\bm{l}^k_{nt'})^{\top}).
\end{equation}
$\alpha_{(nt,nt')}$ is the weight coefficient computed by a softmax function,
and takes into account of the correlation between $(n,t)$ and $(n,t')$ which consists of same word but different frames.
Next, we transform $\widehat{\bm{l}}^v_{nt}$ back to the same dimension as $\bm{f}_{nt}$ via a linear layer and add it element-wise with $\bm{f}_{nt}$ to form a residual connection \cite{he2016deep}:
\begin{equation}
    \widehat{\bm{f}}_{nt} =  \widehat{\bm{l}}^v_{nt}\bm{W}^b + \bm{f}_{nt}.
\end{equation}
We use $\widehat{\bm{F}}_n= \{\widehat{\bm{f}}_{nt} \}^T_{t=1} \in \mathbb{R}^{T\times 2D}$ to denote the collection of $\widehat{\bm{f}}_{nt}$ at all video frames.
Based on $\widehat{\bm{F}}_n$, we can capture long-range dependencies among the video under the $n$-th word guidance. In this stage, we utilize $N$ such multi-modal self attention modules to capture different word-video dependencies. The final query-guided video feature representation is average pooled over all $N$ words of the query as:
\begin{equation}
    \widehat{\bm{F}} = \text{Average}(\{\widehat{\bm{F}}_n\}_{n=1}^N) = \frac{\sum_{n=1}^N \widehat{\bm{F}}_n}{N}.
\end{equation}

\subsection{Training Details}

To train our CBLN, 
we utilize the scaled Intersection over Union (IoU) values as the supervision signal. Specifically, we compute the IoU score $o_p$ of each segment $p$ with the ground truth, and scale $o_p$ as the supervision signal by:
\begin{equation}
    o_p = o_p / \text{Max}(\bm{O}),
\end{equation}
where $\text{Max}(\bm{O})$ denotes the maximum IoU score among all IoU scores $\bm{O}=\{o_p\}^{T\times T}_{p=1}\in \mathbb{R}^{T\times T}$. Our network is trained by a binary cross entropy loss as follows:
\begin{equation}
    \setlength{\abovedisplayskip}{1pt}
    \setlength{\belowdisplayskip}{1pt}
    \mathcal{L}= -\frac{1}{T\times T}\sum_{p=1}^{T\times T}o_p\text{log}(\bm{M}_p)+(1-o_p)\text{log}(1-\bm{M}_p),
\end{equation}
where $\bm{M}_p$ is the score of the segment $p$ in the output $\bm{M}$.


\section{Experiments}

\subsection{Datasets and Evaluation}
\noindent \textbf{ActivityNet Captions.}
ActivityNet Captions \cite{krishna2017dense} contains 20000 untrimmed videos with 100000 descriptions from YouTube. The videos are 2 minutes on average, and the annotated video clips have much larger variation, ranging from several seconds to over 3 minutes. Following public split, we use 37417, 17505, and 17031 sentence-video pairs for training, validation, and testing respectively.

\noindent \textbf{TACoS.}
TACoS \cite{regneri2013grounding} is widely used on TSG task and contain 127 videos. The videos from TACoS are collected from cooking scenarios, thus lacking the diversity. They are around 7 minutes on average. We use the same split as \cite{gao2017tall}, which includes 10146, 4589, 4083 query-segment pairs for training, validation and testing.

\noindent \textbf{Charades-STA.}
Charades-STA is built on the Charades dataset \cite{sigurdsson2016hollywood}, which focuses on indoor activities. 
In total, the video length on the Charades-STA dataset is 30 seconds on average, and there are 12408 and 3720 moment-query pairs in the training and testing sets, respectively.

\noindent \textbf{Evaluation.}
Following previous works \cite{gao2017tall,zhang2019learning,zeng2020dense}, we adopt “R@n, IoU=m” as our evaluation metrics. The “R@n, IoU=m” is defined as the percentage of at least one of top-n selected moments having IoU larger than m.

\subsection{Implementation Details}
For video encoding, we apply C3D \cite{tran2015learning} to encode the videos on all three datasets, and also extract the I3D \cite{carreira2017quo} and VGG \cite{simonyan2014very} features on Charades-STA dataset.
Since some videos are overlong, we set the length of video feature sequences to 200 for ActivityNet Captions and TACoS datasets, 64 for Charades-STA dataset, respectively. As for sentence encoding, we utilize Glove word2vec \cite{pennington2014glove} to embed each word to 300 dimension features. The hidden state dimensions of bi-directional GRU and BiLSTM are set to 512. 
We train our model with an Adam optimizer with leaning rate $8 \times 10^{-4}$, $3 \times 10^{-4}$, $4 \times 10^{-4}$ for ActivityNet Captions, TACoS, and Charades-STA datasets, respectively. The batch size is set to 64. More model details can be found in our supplementary material.

\begin{table}[t!]
    \small
    \centering
    \caption{Comparisons on ActivityNet using C3D features.}
    \scalebox{0.92}{\setlength{\tabcolsep}{0.5mm}{
    \begin{tabular}{c|cccccc}
    \hline
    \multirow{2}*{Method} 
    & R@1, & R@1, & R@1, & R@5, & R@5, & R@5,\\ 
    ~ & IoU=0.3 & IoU=0.5 & IoU=0.7 & IoU=0.3 & IoU=0.5 & IoU=0.7 \\ \hline \hline
    TGN \cite{chen2018temporally} & 45.51 & 28.47 & - & 57.32 & 43.33 & - \\
    CTRL \cite{gao2017tall} & 47.43 & 29.01 & 10.34 & 75.32 & 59.17 & 37.54  \\
    ACRN \cite{liu2018attentive} & 49.70 & 31.67 & 11.25 & 76.50 & 60.34 & 38.57  \\
    QSPN \cite{xu2019multilevel} & 52.13 & 33.26 & 13.43 & 77.72 & 62.39 & 40.78 \\
    CBP \cite{wang2019temporally} & 54.30 & 35.76 & 17.80 & 77.63 & 65.89 & 46.20 \\
    SCDM \cite{yuan2019semantic} & 54.80 & 36.75 & 19.86 & 77.29 & 64.99 & 41.53 \\
    LGI \cite{mun2020local} & 58.52 & 41.51 & 23.07 & - & - & - \\
    2D-TAN \cite{zhang2019learning} & 59.45 & 44.51 & 26.54 & 85.53 & 77.13 & 61.96  \\
    CMIN \cite{zhang2019cross} & 63.61 & 43.40 & 23.88 & 80.54 & 67.95 & 50.73 \\ 
    DRN \cite{zeng2020dense} & - & 45.45 & 24.36 & - & 77.97 & 50.30 \\ \hline
    Ours & \textbf{66.34} & \textbf{48.12} & \textbf{27.60} & \textbf{88.91} & \textbf{79.32} & \textbf{63.41} \\ \hline
    \end{tabular}}}
    \label{tab:activity}
\end{table}

\begin{table}[t!]
    \small
    \centering
    \caption{Comparisons on TACoS using C3D features.}
    \scalebox{0.92}{\setlength{\tabcolsep}{0.5mm}{
    \begin{tabular}{c|ccccccccccccc}
    \hline
    \multirow{2}*{Method} & R@1, & R@1, & R@1, & R@5, & R@5, & R@5,  \\ 
    ~ & IoU=0.1 & IoU=0.3 & IoU=0.5 & IoU=0.1 & IoU=0.3 & IoU=0.5  \\ \hline \hline
    ACRN \cite{liu2018attentive} & 24.22 & 19.52 & 14.62 & 47.42 & 34.97 & 24.88 \\
    CTRL \cite{gao2017tall} & 24.32 & 18.32 & 13.30 & 48.73 & 36.69 & 25.42 \\
    QSPN \cite{xu2019multilevel} & 25.31 & 20.15 & 15.23 & 53.21 & 36.72 & 25.30\\
    CMIN \cite{zhang2019cross} & 32.48 & 24.64 & 18.05 & 62.13 & 38.46 & 27.02\\
    SCDM \cite{yuan2019semantic} &  - & 26.11 & 21.17 & - & 40.16 & 32.18\\
    CBP \cite{wang2019temporally} & - & 27.31 & 24.79 & - & 43.64 & 37.40 \\
    TGN \cite{chen2018temporally} & 41.87 & 21.77 & 18.90 & 53.40 & 39.06 & 31.02\\
    DRN \cite{zeng2020dense} & - & - & 23.17 & - & - & 33.36 \\
    2D-TAN \cite{zhang2019learning} & 47.59 & 37.29 & 25.32 & 70.31 & 57.81 & 45.04 \\
     \hline
    Ours & \textbf{49.16} & \textbf{38.98} & \textbf{27.65} & \textbf{73.12} & \textbf{59.96} & \textbf{46.24} \\ \hline
    \end{tabular}}}
    \label{tab:tacos}
    \vspace{-10pt}
\end{table}

\subsection{Comparisons with state-of-the-arts}

\noindent \textbf{Comparisons on ActivityNet Captions.}
We compare our CBLN with the state-of-the-art methods on the ActivityNet Captions dataset in Table \ref{tab:activity}. We follow the previous methods to use C3D features for fair comparisons. Particularly, our model outperforms the previously best method DRN \cite{zeng2020dense} by
3.24\% and 13.11\% absolute improvement in terms of R@1, IoU=0.7 and R@5, IoU=0.7, respectively. Compared to the method 2D-TAN \cite{zhang2019learning}, we also outperform them by 6.89\%, 3.61\%, 1.06\%, 3.38\%, 2.19\% and 1.45\% in terms of all metrics, respectively. 

\noindent \textbf{Comparisons on TACoS.}
We compare our CBLN with the state of-the-art methods with the same C3D features in Table \ref{tab:tacos}. On TACoS dataset, the cooking activities take place in the same kitchen scene with slightly varied cooking objects, thus showing the challenging nature of this dataset. Despite its difficulty, our model still reaches highest scores in terms of both R@1 and R@5 when IoU=0.5, and outperforms both 2D-TAN \cite{zhang2019learning} and DRN \cite{zeng2020dense} by a great margin.


\noindent \textbf{Comparisons on Charades-STA.}
Table \ref{tab:charades} reports the grounding results of various methods. Our CBLN reaches the highest results over all evaluation metrics. Specifically, when using the same VGG features, compared to the previously best method 2D-TAN \cite{zhang2019learning}, our model brings the absolute improvement of 3.86\%, 1.19\%, 9.06\% and 5.34\% on all metrics, respectively. For fair comparisons with GDP~\cite{chenrethinking} and LGI~\cite{mun2020local}, we also perform experiments with same features (i.e., C3D and I3D) reported in their papers. It is obvious that our model still performs better. All these results again verify the effectiveness of our model.

\subsection{Ablation Studies}
In this section, we will perform in-depth ablation studies to evaluate the effect of each component in our CBLN on ActivityNet Captions dataset.

\begin{table}[t!]
    \small
    \centering
    \caption{Comparisons with state-of-the-arts on Charades-STA.}
    \setlength{\tabcolsep}{0.5mm}{
    \begin{tabular}{c|c|cccc}
    \hline
    \multirow{2}*{Method} & \multirow{2}*{Feature} & R@1, & R@1, & R@5, & R@5,  \\ 
    ~ & ~ & IoU=0.5 & IoU=0.7 & IoU=0.5 & IoU=0.7 \\ \hline \hline
    SAP \cite{chen2019semantic} & VGG & 27.42 & 13.36 & 66.37 & 38.15 \\
    2D-TAN \cite{zhang2019learning} & VGG & 39.81 & 23.25 & 79.33 & 51.15 \\ 
    Ours & VGG & \textbf{43.67} & \textbf{24.44} & \textbf{88.39} & \textbf{56.49} \\ \hline \hline
    CTRL \cite{gao2017tall} & C3D & 23.63 & 8.89 & 58.92 & 29.57 \\
    QSPN \cite{xu2019multilevel} & C3D & 35.60 & 15.80 & 79.40 & 45.40\\
    CBP \cite{wang2019temporally} & C3D & 36.80 & 18.87 & 70.94 & 50.19\\
    GDP \cite{chenrethinking} & C3D & 39.47 & 18.49 & - & -\\
    Ours & C3D & \textbf{47.94} & \textbf{28.22} & \textbf{88.20} & \textbf{57.47} \\ \hline \hline
    DRN \cite{zeng2020dense} & I3D & 53.09 & 31.75 & 89.06 & 60.05 \\
    SCDM \cite{yuan2019semantic} & I3D & 54.44 & 33.43 & 74.43 & 58.08\\ 
    LGI \cite{mun2020local} & I3D & 59.46 & 35.48 & - & - \\
    Ours & I3D & \textbf{61.13} & \textbf{38.22} & \textbf{90.33} & \textbf{61.69} \\ \hline
    \end{tabular}}
    \label{tab:charades}
\end{table}

\begin{table}[t!]
    \small
    \centering
    \caption{Ablation studies of the baseline model with different grounding strategies on ActivityNet Captions, where * denote the baseline model with biaffine mechanism.}
    \begin{tabular}{lcc|cc}
    \hline
    \multirow{2}*{Model} & ~& ~ & R@1, & R@5, \\ 
    ~ & ~& ~ & IoU=0.7 & IoU=0.7 \\ \hline \hline
    Baseline (regression) &~ & ~& 17.02 & 45.83 \\
    Baseline (proposal-match) &~ & ~& 20.10 & 51.96 \\
    Baseline* (biaffine) &~& ~ & \textbf{22.74} & \textbf{53.79} \\ \hline
    \end{tabular}
    \label{tab:baseline}
    \vspace{-10pt}
\end{table}


\noindent \textbf{Main ablation studies.} First of all, to investigate the effectiveness of biaffine localization module in this paper, we build up three baseline models with different grounding heads. For a fair comparison, we keep video/query encoders and cross-modal attention mechanism \cite{mithun2019weakly,zhang2019cross} consistent in all baselines. Baseline (regression) directly regresses temporal boundary \cite{yuan2019find,mun2020local}, while Baseline (proposal-match) designs pre-defined proposals to match the query \cite{yuan2019semantic,zhang2019cross}.  
Table \ref{tab:baseline} shows that the Baseline* (biaffine) with our proposed biaffine mechanism significantly outperforms the other two baselines. It demonstrates that the biaffine mechanism is more suitable for segment localization in TSG task as it can learn detailed start-end frames interaction and get rid of handcrafted proposals compared to previous methods. 

Next, to investigate the contribution of the proposed multi-modal self attention (MMSA) module and multi-context biaffine localization (MCBL) module, we also implement three variants of our model as shown in Table \ref{tab:ablation1}. Compared to the baseline*, MMSA captures more fine-grained query-video interactions and outperforms it by 1.48\% and 3.77\% in R@1, and R@5,IoU=0.7, respectively. Moreover, MCBL brings the highest improvement on both two metrics (i.e. 3.74\% and 7.19\%), which demonstrates its effectiveness of aggregating local-global contexts.

\begin{table}[t!]
    \small
    \centering
    \caption{Results of main ablation studies on ActivityNet Captions.}
    \setlength{\tabcolsep}{0.5mm}{
    \begin{tabular}{cc|cc}
    \hline
    \multirow{2}*{\tabincell{c}{Multi-modal self\\attention (MMSA)}} & \multirow{2}*{\tabincell{c}{Multi-context biaffine\\localization (MCBL)}} & R@1, & R@5, \\ 
    ~ & ~ & IoU=0.7 & IoU=0.7 \\ \hline \hline
    $\times$ & $\times$ & 22.74 & 53.79 \\
    \checkmark & $\times$ & 24.22 & 57.56 \\
    $\times$ & \checkmark & 26.48 & 60.98 \\
   \checkmark & \checkmark & \textbf{27.60} & \textbf{63.41} \\
    \hline
    \end{tabular}}
    \label{tab:ablation1}
\end{table}

\begin{table}[t!]
    \small
    \centering
    \caption{Performance comparison with varying different local-global contexts in MCBL module on ActivityNet Captions dataset.}
    \begin{tabular}{c|c|cc}
    \hline
    \multirow{2}*{Component} & \multirow{2}*{Changes} & R@1, & R@5, \\ 
    ~ & ~ & IoU=0.7 & IoU=0.7 \\ \hline \hline
    \multirow{3}*{\tabincell{c}{local\\extraction}} & mean-pooling & 25.09 & 59.98 \\
   ~ & max-pooling & 25.47 & 61.29 \\
   ~ & concatenate & \textbf{27.60} & \textbf{63.41} \\ \hline
   \multirow{3}*{\tabincell{c}{global\\spanning}} & sampling & 24.88 & 58.25 \\
   ~ & mean-pooling & 26.02 & 61.83 \\
   ~ & max-pooling & \textbf{27.60} & \textbf{63.41} \\ \hline
   \multirow{8}*{\tabincell{c}{local\\scale $K^l$}} & \{1\} & 25.45 & 61.12 \\
   ~ & \{3\} & 25.72 & 61.52 \\
   ~ & \{5\} & 25.63 & 61.49  \\
   ~ & \{7\} & 25.17 & 60.96 \\
   ~ & \{1,3\} & 26.76 & 62.17 \\
   ~ & \{3,5,7\} & 27.13 & 62.63 \\ 
   ~ & \{1,3,5\} & \textbf{27.60} & 63.41\\
   ~ & \{1,3,5,7\} & 27.38 & \textbf{63.49} \\ 
   \hline
   \multirow{8}*{\tabincell{c}{global\\scale $K^g$}} & \{1\} & 25.41 & 61.34 \\
   ~ & \{2\} & 25.13 & 60.99\\
   ~ & \{4\} & 25.06 & 60.75\\
   ~ & \{8\} & 24.89 & 60.38 \\
   ~ & \{1,2\} & 26.48 & 62.02 \\
   ~ & \{2,4,8\} & 26.91 & 62.77 \\
   ~ & \{1,2,4\} & 27.60 & 63.41 \\
   ~ & \{1,2,4,8\} & \textbf{27.72} & \textbf{63.68} \\ 
   \hline
    \end{tabular}
    \label{tab:ablation2}
    \vspace{-10pt}
\end{table}

\noindent \textbf{Analysis on local-global contexts generation.}
As shown in Table \ref{tab:ablation2}, we conduct the investigation on the impact of different local-global contexts modeling in multi-context biaffine localization (MCBL) module. First, 
for local context extraction, we need to preserve the details of each adjacent frame, thus the pooling strategy may lose some discriminative information contained in each frame features. The results also show that the concatenation performs better than both mean- and max-pooling. 
For global context spanning, it is difficult to work with all the raw features. Therefore, we select representative features by sub-sampling or pooling. In our experiments, we can find that max-pooling is superior to both random sampling and mean-pooling, as random sampling may lose discriminative frame feature and mean-pooling smooths away salient features that are otherwise preserved by max-pooling. 

Besides, we also show the influence on different scales of both local $K^l$ and global $K^g$ window sizes. With more kinds of different scales, the model usually performs better than individual scale. For local scale $K^l$, we choose $K^g=\{1,3,5\}$. 
As for global scale $K^g$, the variant with four kinds of scales $\{1,2,4,8\}$ achieves the best result but only performs marginally better than the three-scales one $\{1,2,4\}$ at the expense of significantly larger cost of GPU memory. Thus, we choose $K^g=\{1,2,4\}$ for global contexts in our experiments.

\begin{table}[t!]
    \small
    \centering
    \caption{Performance comparison with varying combinations
    of modules in local-global aggregation on ActivityNet Captions.}
    \begin{tabular}{l|cc}
    \hline
    \multirow{2}*{Changes} & R@1 & R@5 \\ 
    ~ & IoU=0.7 & IoU=0.7 \\ \hline \hline
    Full model & \textbf{27.60} & \textbf{63.41} \\ \hline
    remove all non-local blocks & 25.05 & 60.88 \\
    replace non-local with linear & 25.93 & 61.74\\
    replace final linear with pooling & 26.86 & 62.78\\
    \hline
    \end{tabular}
    \label{tab:ablation3}
\end{table} 

\begin{table}[t!]
    \small
    \centering
    \caption{Performance comparison with varying fusion strategies on multiple information on ActivityNet Captions.}
    \begin{tabular}{c|c|cc}
    \hline
    \multirow{2}*{Module} & \multirow{2}*{Fusion} & R@1, & R@5, \\ 
    ~ & ~ & IoU=0.7 & IoU=0.7 \\ \hline \hline
    \multirow{3}*{MMSA} & concat+linear & 26.84 & 62.37 \\
    ~ & max-pooling & 25.53 & 61.24 \\
    ~ & mean-pooling & \textbf{27.60} & \textbf{63.41} \\ \hline
    \multirow{2}*{MCBL} & max-pooling & 27.46 & 63.12  \\
    ~ & mean-pooling & \textbf{27.60} & \textbf{63.41} \\ \hline
    \end{tabular}
    \label{tab:ablation4}
\end{table}

\begin{table}[t!]
    \small
    \centering
    \caption{Complexity comparison on ActivityNet.
    ``Speed" denotes the average time to localize one sentence in a given video.}
    \scalebox{0.9}{
    \setlength{\tabcolsep}{1.2mm}{
    \begin{tabular}{cc|ccc}
    \hline
    MMSA & MCBL & R@1, IoU=0.7 & Speed & Memory (batchsize) \\ \hline
    $\checkmark$ & $\checkmark$ & \textbf{27.60} & 0.18s & 10184M (64) \\ 
    $\times$ & $\checkmark$ & 26.48 & 0.14s & 8747M (64)\\
    $\checkmark$ & $\times$ & 24.22 & 0.09s & 5898M (64)\\
    \hline
    \multicolumn{2}{c|}{CMIN} & 23.88 & 0.08s & 5692M (64) \\
    \multicolumn{2}{c|}{2D-TAN} & 26.54 & 0.57s & 10572M (16)\\ \hline
    \end{tabular}}}
    \label{tab:ablation5}
    \vspace{-10pt}
\end{table}


\noindent \textbf{Analysis on local-global contexts aggregation.}
To interact information of both local and global contexts, we apply stacked non-local blocks to combine them in a learnable way in the local-global aggregation module of MCBL. As shown in Table \ref{tab:ablation3}, we can observe that when removing all non-local blocks and separately passing global and local contexts to latter computation, there is a performance drop of 2.55\% and 2.53\% in R@1, and R@5, IoU=0.7, respectively. When we replace non-local block with concatenation and a linear layer, there is also a drop of 1.73\% and 1.67\%. These results demonstrate that the stacked non-local blocks are effective for local-global contexts re-weighting. We also evaluate the performance when we replace the final linear layer (for multiple local/global combination, as shown in Figure \ref{fig:biaffine}) with max-pooling, it drops 0.74\% and 0.63\%.

\noindent \textbf{Analysis on the fusion strategies.} To better fuse the multiple information obtained by the outputs of both multi-modal self attention (MMSA) module and multi-context biaffine localization (MCBL) module, we compare different fusion operations as shown in Table \ref{tab:ablation4}. We can find that mean-pooling achieves the best performance in both two modules.

\noindent \textbf{Analysis on model Complexity.} To investigate the complexity of our model, we give an in-depth study in terms of speed and memory as shown in Table \ref{tab:ablation5}. Though our full model is not more efficient than CMIN, it outperforms CMIN with a large margin. Our w/o MCBL method still achieves better performance with similar computational cost (compared to CMIN). Compared to 2D-TAN, our full model performs better and much more efficient.

\subsection{Qualitative Results}
\begin{figure}[t]
\centering
\includegraphics[width=0.47\textwidth]{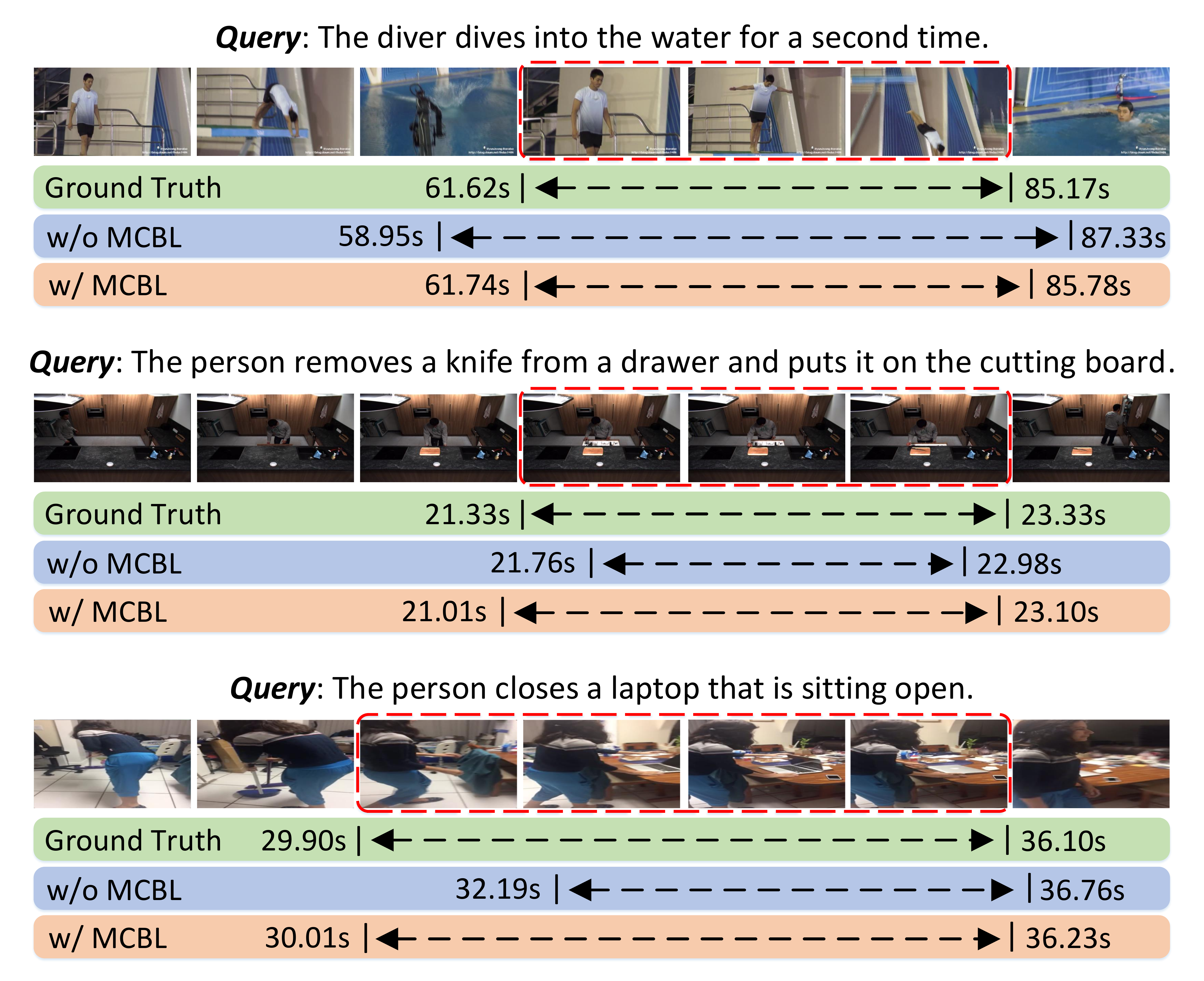}
\caption{Qualitative results on ActivityNet Captions (top), TACoS (middle), and Charades-STA (bottom) datasets, respectively.}
\label{fig:result}
\vspace{-10pt}
\end{figure}

Figure \ref{fig:result} shows some qualitative results from three datasets. Our multi-context biaffine localization module can provide more contextual details about the segment, thus achieves better grounding results.

\vspace{-5pt}
\section{Conclusion}
In this paper, we have proposed a novel context-aware biaffine localizing network, called CBLN, for temporal sentence grounding. The key to CBLN is that we reformulate this task from a new perspective for scoring all pairs of start and end indices simultaneously by a biaffine mechanism. To enrich the feature representation of each start/end frame, we additionally integrate them with multi-scale local and global contexts. A multi-modal self attention module is also developed to generate fine-grained query-guided video representation for such biaffine-based strategy. The experiments on three public datasets demonstrate the performance of CBLN, which brings significant improvements over the state-of-the-art methods.

\noindent \textbf{Acknowledgements.} This work was supported in part by the National Natural Science Foundation of China (No.61972448, No.61902347), and the Zhejiang Provincial Natural Science Foundation (No. LQ19F020002).

{\small
\bibliographystyle{ieee_fullname}
\bibliography{egbib}
}

\end{document}